\documentclass[runningheads]{llncs}

\usepackage{eccv}
\usepackage[dvipsnames]{xcolor}
\usepackage{comment}

\usepackage{eccvabbrv}
\usepackage{graphicx}
\usepackage{booktabs}
\usepackage[accsupp]{axessibility} 
\usepackage{hyperref}
\usepackage{orcidlink}

\newcommand{\contextimgHuman}{I_c^h} % CONTEXT image with human
\newcommand{\contextimg}{\mathcal{I}} % CONTEXT image
\newcommand{\poseinput}{\mathcal{S}} % Input pose heatmaps
\newcommand{\contextimgHumanmask}{M_h} % CONTEXT image mask of the human
\newcommand{\contextimgBackmask}{M_b} % CONTEXT image mask of the human
\newcommand{\contextimgDown}{i_c} % CONTEXT image
\newcommand{\contextimgVector}{\mathcal{C}} % CONTEXT image
\renewcommand{\vec}[1]{\mathbf{#1}}
\newcommand{\weight}{\lambda} % loss weight parameter - lambda or 
\newcommand{\lV}{L_{V}} % Loss D superres
\newcommand{\lK}{D_{KL}} % Loss DKL
\newcommand{\expec}{\mathbb{E}}

\newcommand{\model}{\epsilon_\theta}

\newcommand{\encoder}{\mathcal{E}}

\begin{document}

% ---------------------------------------------------------------
\def\OURS{Environment-Specific People}
\title{Synthesizing \underline{E}nvironment-\underline{S}pecific \underline{P}eople in Photographs} 

\author{Mirela Ostrek\inst{1}\orcidlink{0009-0009-9987-646X} \and
Carol O'Sullivan\inst{2}\orcidlink{0000-0003-3772-4961} \and
Michael J. Black\inst{1}\orcidlink{0000-0001-6077-4540} \and
Justus Thies\inst{1,3}\orcidlink{0000-0002-0056-9825}}

\authorrunning{M. Ostrek et al.}

\institute{Max Planck Institute for Intelligent Systems, Tübingen, Germany \and
Trinity College Dublin, Dublin, Ireland \and
Technical University of Darmstadt, Darmstadt, Germany
}

\sloppy

\newcommand{\teaserCaption}{
        \textbf{\textit{ESP} generates people in semantically appropriate, environment-inspired clothing.} Given a 2D pose and an environment (top), people are inpainted into the photo based on a context-aware generative model 
        (sample 1, 2, 3).
        }

\renewcommand\twocolumn[1][]{#1}%
\maketitle
\begin{center}
    \centering
    \captionsetup{type=figure}
    \includegraphics[width=\textwidth]{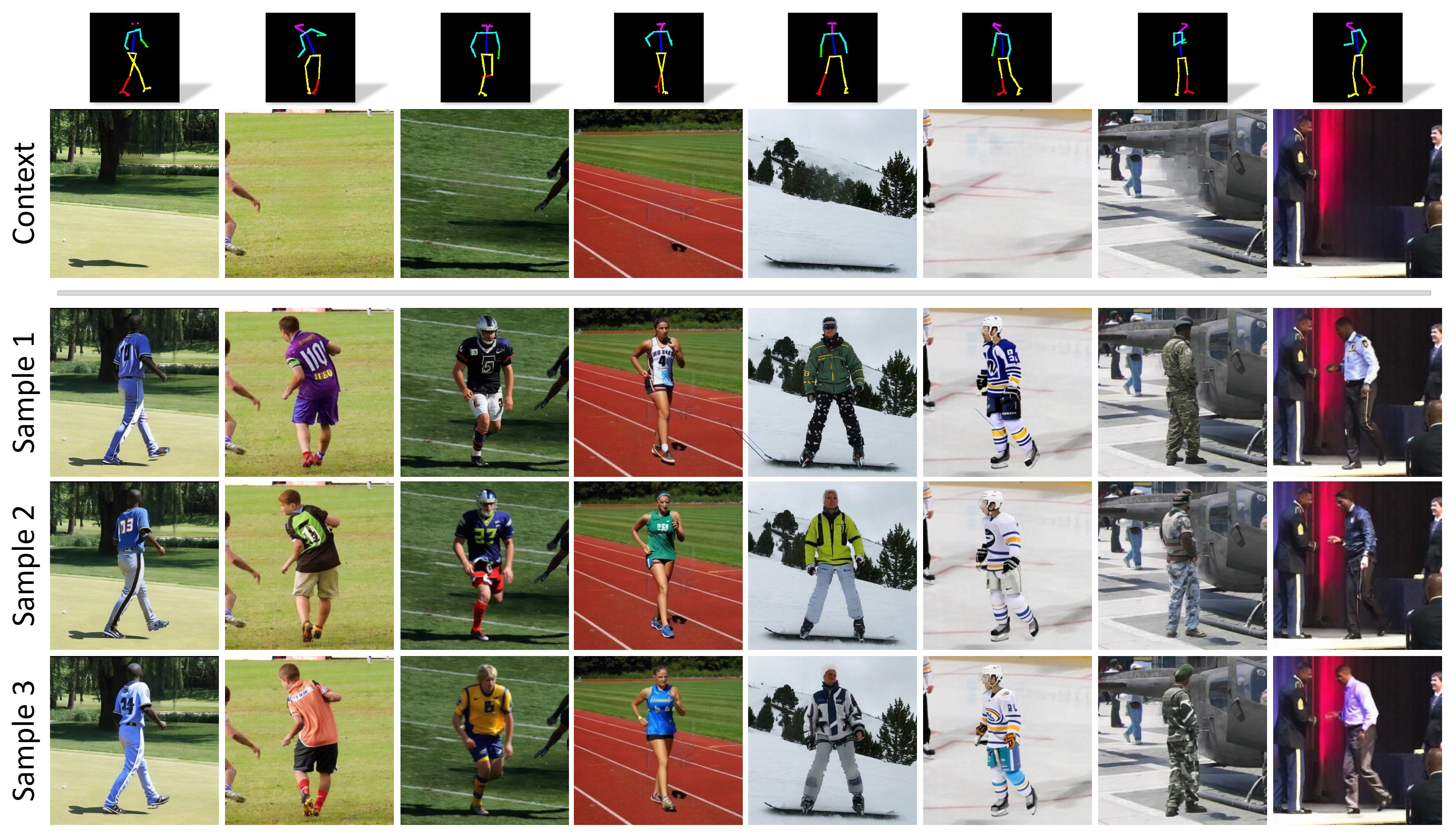}
    \caption{\teaserCaption}
    \label{fig:teaser}
\end{center}

\begin{abstract}
We present \textit{ESP}, a novel method for context-aware full-body generation, that enables photo-realistic synthesis and inpainting of people wearing clothing that is semantically appropriate for the scene depicted in an input photograph.
\textit{ESP} is conditioned on a 2D pose and contextual cues that are extracted from the photograph of the scene and integrated into the generation process, where the clothing is modeled explicitly with human parsing masks (HPM).
Generated HPMs are used as tight guiding masks for inpainting, such that no changes are made to the original background.
Our models are trained on a dataset containing a set of in-the-wild photographs of people covering a wide range of different environments.
The method is analyzed quantitatively and qualitatively, and we show that \textit{ESP} outperforms the state-of-the-art on the task of contextual full-body generation.
\href{https://esp.is.tue.mpg.de/}{https://esp.is.tue.mpg.de/}
\end{abstract}    
%
%\clearpage
\section{Introduction}
\label{sec:introduction}
In recent years, great progress has been made on all aspects of image generation and editing.
GAN-based methods~\cite{zhao2020uctgan, zheng2019pluralistic, teterwak2019boundless, li2021controllable, zhao2021large, zhao2021prior} allow for photo-realistic image synthesis on well-crafted datasets, and diffusion-based models~\cite{rombach2021highresolution, zhang2023adding, lugmayr2022repaint} make it possible for the original image content to be seamlessly removed or replaced via inpainting.
Rapid advancements in generative AI have resulted in tools that take text prompts and produce photo-realistic images of humans \cite{dalle-2,geminiteam2023gemini}.
Constraining the generation requires crafting text prompts to achieve the desired result.
In this paper, our goal is to take an image of a scene and a 2D pose and generate a realistic person, in the desired pose, wearing scene-appropriate clothing; no text prompts are required.
For example, if the context of the depicted scene is a football field, our method generates people dressed in appropriate athletic wear such as shorts, a jersey, football helmet, etc.; see Figure \ref{fig:teaser}. 
However, if the context switches to snowy mountains, the inserted person will be wearing a winter jacket, ice skates, or ski attire.
One line of work in full-body generation is centered around using StyleGAN~\cite{fruhstuck2022insetgan, fu2022stylegan} to generate full-body samples. 
Such methods are trained on high-quality data, such as DeepFashion~\cite{liuLQWTcvpr16DeepFashion}, that is curated and hand annotated.
To reduce its high dimensionality, data is first normalized and filtered.
This is a crucial step that is contributing to the success of StyleGAN-based methods.
While the images that are generated in this way are of high-quality, 
the generated clothing is typically not correlated with the scene due to the unnatural settings present in the training data.
For example, in fashion photography, images are usually captured in a studio setting, where the subject is fully lit and the background is very simple, e.g., a white screen. Therefore, there is minimal environmental context available.  
The more recent line of work focuses on diffusion-based methods that generate images from text, e.g.~large pre-trained latent diffusion models (LDM) such as Stable Diffusion~\cite{rombach2021highresolution} and its conditional variations (e.g. ControlNet~\cite{zhang2023adding}).
The main advantage of such methods is the existence of large pre-trained image priors that are usually fine-tuned on human data including full-bodies in the wild.
However, in order to generate sufficiently varied samples, these text-to-image (T2M) models require accurate textual descriptions of the desired output.
In contrast, here we seek an automated solution and do not assume that textual descriptions of the appropriate clothing are available.
Instead, we focus on visual information alone, investigating whether the image context can be used as a sole driving source for the generation of environment-specific people.
To this end, we take advantage of both StyleGAN-based and diffusion-based methods, combining their strengths to devise a purely image-based system for our task.
As part of our pipeline, contextual cues are first extracted from a given environment photo with a VAE and, then, fed into a StyleGAN-based human parsing map (HPM) generator that predicts both clothing semantics and the original environment.
The key insight is that, when combined with a random vector, the extracted contextual cues form our \textbf{contextual-style vectors} that allow for a stochastic generation of a large variety of context-appropriate clothing.
These vectors are connected back to the original environment, as they jointly drive the generation of both the HPMs and the corresponding environment,
making the HPM generation context-aware.
To disentangle the pose from the context, the generator is further conditioned on 2D pose renderings and the network's computational graph is extended through an image-to-image (I2I) translation module (similar to pix2pix~\cite{isola2017image}).
Finally, leveraging a large pre-trained T2M LDM for its state-of-the-art image synthesis and inpainting capacities~\cite{rombach2021highresolution}, generated HPMs are used as input conditioning and guidance during the inpainting and super-resolution process.
Our method is evaluated quantitatively and qualitatively, demonstrating that our approach outperforms the state of the art on the task of contextual full-body generation based on the following contributions:
\begin{itemize}
  \item[$\circ$] A StyleGAN-based method for stochastic generation of context-aware HPMs that are generated using our newly introduced contextual style vectors;
  \item[$\circ$] An extension of the computational graph that is trained end-to-end with the StyleGAN generator which enables pose-guided I2I translation;
  \item[$\circ$] A system that leverages a large pre-trained T2M latent diffusion model (Stable Diffusion), allowing for super-resolution imaging and inpainting of the environment-specific people into the existing photographs.
\end{itemize}
\section{Related work}
\label{sec:related_work}

Generative Adversarial Networks (GAN)~\cite{NIPS2014_5ca3e9b1} have proven to be a very effective tool for synthesizing realistic images.
StyleGAN-based methods~\cite{karras2019style, karras2020training, karras2021alias, brock2018large} have been used to generate photo-realistic imagery of a variety of objects and scenes~\cite{tewari2020state, tewari2022advances}, including humans~\cite{lewis2021tryongan, sarkar2021humangan}.
More recently, large latent diffusion models have been in the spotlight~\cite{po2023state} due to their unparalled image generation capabilities, conditioned on text (T2M) ~\cite{rombach2021highresolution, dalle, dalle-2, midjourney, imagen} and other controls ~\cite{zhang2023adding}.

\paragraph{\textbf{Full-body generation:}}
StyleGAN-based state-of-the-art methods for unconditional full-body generation include InsetGAN~\cite{fruhstuck2022insetgan} and StyleGAN-Human~\cite{fu2022stylegan}.
Both methods use well-prepared datasets that are either proprietary or contain fashion imagery.
InsetGAN composes multiple GANs to handle the complex task of generating high resolution full body human images.
One GAN is trained to generate the global structure, i.e., the body, while the remaining GANs are each trained separately for different local regions, e.g., face or shoes. 
The latent spaces of different GANs are jointly optimized for a seamless combination to synthesize the whole person.
StyleGAN-Human involves training StyleGAN2~\cite{Karras2019stylegan2} on a large, self-curated dataset, consisting of around 230K full body human images.
The background in this data is often white or plain, or the link between the background and the clothing is atypical. 
We use a small dataset of in-the-wild imagery~\cite{zhou2021human} for training, where people are clothed according to the environment, e.g., football gear is worn on a football field and warm winter attire in the snow.
Pose and/or appearance information are used to control the process of conditional full-body generation~\cite{albahar2021pose, ma2017pose, men2020controllable, siarohin2019appearance, sarkar2021style}.
In a two-step process, Lewis et al.~\cite{lewis2021tryongan} first train a dense-pose conditioned StyleGAN2 model with a clothing segmentation branch, after which a new image is synthesized using layered latent space interpolation to transfer a garment to a target body.
Grigorev et al.~\cite{grigorev2021stylepeople} combine 3D parametric mesh-based models of human bodies~\cite{pavlakos2019expressive} with neural textures generated by a modified StyleGAN2 model.
Ren et al.~\cite{ren2020deep} propose a deep architecture, conditioned on human pose, to combine flow-based operations with an attention mechanism. 
This architecture has also been extended for use in an unsupervised setting~\cite{sanyal2021learning}.
Sarkar et al.~\cite{sarkar2021humangan} warp encoded part-based latent appearance vectors in a normalized pose-independent space to generated synthetic people in specific poses.
Our twist considering this line of work is to tackle the problem from a novel perspective, by incorporating image context as a conditioning through the mapping network. 
In this way, we generate environment-specific people that may be inpainted into the existing photographs, while respecting the scene context in terms of the clothing semantics.

\paragraph{\textbf{Inpainting:}}
The task of generating new content in specified regions of an existing image is referred to as inpainting.
This is a well-studied problem that is particularly suitable for our task~\cite{teterwak2019boundless, li2021controllable, zhao2021prior}. 
Motivated by the StyleGAN2 modulated convolutions, Zhao et al.~\cite{zhao2021large} improve inpainting reconstruction by using a comodulation layer.
Approaches that build on a variational auto-encoder based architecture have also been proposed~\cite{zheng2019pluralistic, zhao2020uctgan}.
Lugmayr et al.~\cite{lugmayr2022repaint} improved on the previous methods by harnessing the expressiveness of a pre-trained denoising diffusion probabilistic model (DDPM)~\cite{ho2020denoising},
which had previously shown impressive results in guided image synthesis in pixel space~\cite{choi2021ilvr, meng2021sdedit}.
Using a pre-trained DDPM as a generative prior, they alter the reverse diffusion iterations by sampling the unmasked regions using the given image information.
However, this approach slows down the image generation process of a DDPM.
More recently, similar ideas in the realm of latent diffusion models have been explored using denoising diffusion implicit models (DDIM)~\cite{song2020denoising}, where the denoising steps are skipped for faster inference.
Most recent image generators that have the inpainting capacity use this principle and we take advantage of these models ~\cite{rombach2021highresolution}.

\paragraph{\textbf{Humans in scenes:}}
Brooks et al.~\cite{brooks2022hallucinating} generate both the scene and a pose-compatible human that matches the scene context at $128 \times 128$, where the context embeddings are calculated from a 2D pose vector and a noise vector. 
In their work, the focus is on generating both compatible scenes and humans inside, conditioned on 2D pose embeddings.  
In contrast, our focus is on inpainting environment-specific people into existing scenes and we calculate context embeddings from the scene context directly.
Kulal et al.~\cite{kulal2023putting} filter a large-scale video dataset (2.4 million clips) at $256 \times 256$ and use it to train a multi-faceted LDM, generating pose-compatible people conditioned on a photo of a reference person and a masked scene. 
The primary focus of this work is on synthesizing plausible poses given a scene context. 
For this purpose, two random frames that contain the same human from each video are used during training.
At test time, the reference person conditioning may be removed, allowing for the synthesis of scene-compatible humans, given scene information.
In contrast, we condition the generation process on a full-body 2D pose rendering (similar to ~\cite{chan2019everybody}) and investigate the connection between the scene and its impact on clothing alone, given a fixed 2D pose.
Additionally, we only use a small dataset containing approx. 23K images at $200 \times 200$, and our results are generated at $512 \times 512$.
Image quality of our method is on par with state-of-the-art image generators including ControlNet~\cite{zhang2023adding} and Stable Diffusion~\cite{rombach2021highresolution} that many other recent high-quality full-body generation methods build on top of ~\cite{liu2023hyperhuman}.

\begin{figure*}[t!]
    \includegraphics[width=\textwidth]{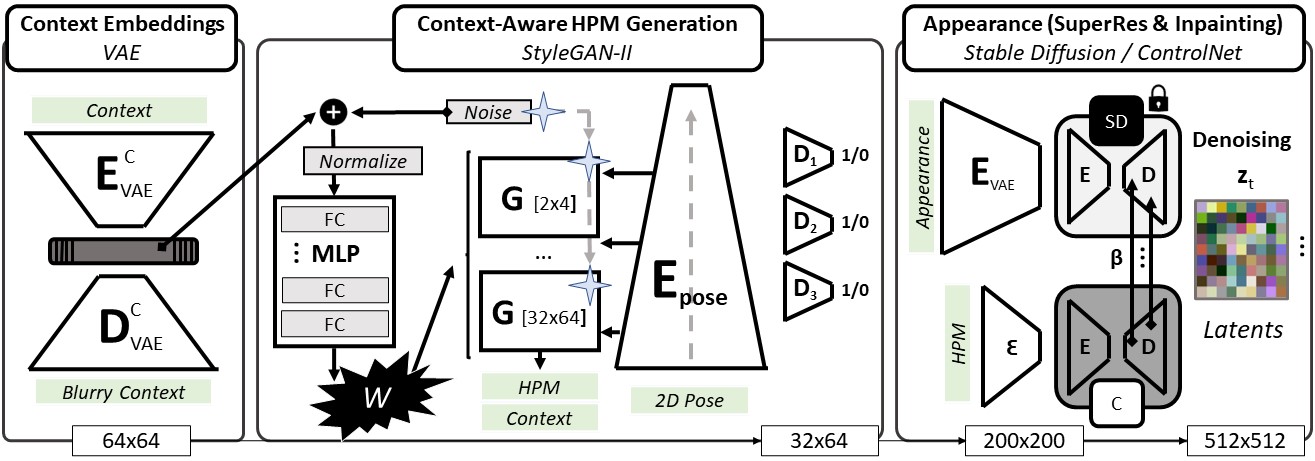}
    \caption{\textbf{System overview:} (I) an input context image is encoded into the latent space of a VAE, giving context embeddings; (II) the latter is then fed, alongside a random vector, creating a contextual style vector, into a pose-conditioned StyleGAN-II HPM generator; (III) generated HPMs are used as input for pretrained Stable Diffusion/CN modules to achieve fine-grained control over the generated clothing during inpainting.
    }
    \label{fig:pipeline}
\end{figure*}

\section{Method}
\label{sec:method}

Our method is shown in \Cref{fig:pipeline}.
First, context embeddings are extracted from image $\mathcal{I}$ that contains the environmental context $\mathcal{C}$.
Then, using the extracted embeddings, human parsing maps (HPMs) $\mathcal{M}$ are generated conditioned on a 2D skeleton rendering $\mathcal{S}$ ($32 \times 64$).
The generated HPMs are then used, with the help of our HPM translation module, as an additional guiding signal for inpainting of environment-specific people into the existing photographs.
In this final step, we take advantage of a large pre-trained LDM (Stable Diffusion for inpainting), allowing for the generation of images at high resolution ($512 \times 512$).

%%%%%%%%%%%%%%%%%%%%%%%%%%%%%%%%%%%%%%%%%%%%%%%%%%%%%%%%%%%%

\subsection{Context embeddings}
We must first find a representation of $\contextimg$ that can be used as the context conditioning to drive the generation of the clothing semantics in the following steps.
During training, we have access to the context image containing the human $\contextimgHuman$ and a corresponding mask of the human $\contextimgHumanmask$. 
We dilate the mask with morphological operations and invert it to get the soft background mask $\contextimgBackmask$.
To remove the foreground human, we apply a pixel-wise mask on the input image, producing the masked image $\contextimg = \contextimgHuman \odot \contextimgBackmask$ .
We choose the compact and normalized latent space of a Variational Auto-Encoder (VAE)~\cite{kingma2013auto,gulrajani2017pixelvae}  to reflect the global structure, illumination and color tone of the context image $\contextimg$.
We train the VAE with $\contextimgDown$, where $\contextimgDown$ is a down-sampled $64 \times 64$ version of $\contextimg$ with the loss:
\begin{equation}
    \begin{split}
        \lV = \weight_1\ ||\contextimgDown - \Tilde{i}_c ||_1 +
        \weight_2 \lK \ ({\scriptstyle \mathcal{N} (\mu_c, \sigma^{2}_c)} \ || \ {\scriptstyle \mathcal{N}(0,1)}) ,
    \end{split}
    \label{eq:vae_loss}
\end{equation}
where $\Tilde{i}_c$ is the reconstructed image from the VAE; 
$\weight$s are the corresponding weights for the different losses;
$\lK$ is the KL divergence loss~\cite{kingma2013auto};
$\mathcal{N}(x,y)$ is the normal distribution with mean $x$ and variance $y$; and
$\mu_c$, $\sigma^{2}_c$ are the mean and variance of the context image.
Once trained, the latent vector $\contextimgVector$ sampled from the VAE for the corresponding context image $\contextimg$ is used for conditioning in the following steps.
Note that the reconstruction loss term is applied only to the regions where humans are present, and, as such, we do not require $\contextimg$ to be computed from $\contextimgHuman$ at inference time.
%

%%%%%%%%%%%%%%%%%%%%%%%%%%%%%%%%%%%%%%%%%%%%%%%%%%%%%%%%%%%%

\subsection{Contextualized generation of HPMs}
We train a StyleGAN2 generator, conditioned on $\contextimgVector$, to synthesize a human parsing map (HPM) and a low-resolution reconstruction of the corresponding context image, for a given 2D pose rendering $\poseinput$.
A $512$-dimensional random vector $z_s$, sampled from $\mathcal{N}(0,1)$, is concatenated with $\contextimgVector$ (also $512$-d) and then passed as input to the StyleGAN2~\cite{Karras2019stylegan2} generator. 
After $z_s$ and $\contextimgVector$ are passed through the mapping network, they are jointly referred to as the \textit{contextual style vector} in this paper.
Concatenating $z_s$ with $\contextimgVector$ is important because there may be more than one outfit that is appropriate for a given environment.
$z_s$ captures the one to many mapping between the context and the semantics.

The HPM generator is additionally conditioned with $\poseinput$ in the image space to enforce the generation of full-body humans in the given pose.
This helps us isolate the pose later on in the experiments and focus on exploring the impact that the environment by itself has on the clothing semantics.
Motivated by Isola et al.~\cite{isola2017image}, to inject the pose conditioning into the generator, we concatenate the input of different style blocks of the generator with the output of the corresponding layers of a pose encoder.
The input to the encoder is an RGB image $\poseinput$ and its layers are designed so that the number of the feature maps and their dimensions match the StyleGAN2 generator ($32 \times 64$).
The generator and the pose encoder are trained end-to-end.
The HPM generator jointly outputs a HPM $\tilde{\mathcal{M}}$ (3 channels) and a blurry background image ($3$ channels).
Note that the blurry background reconstruction from the VAE that is used as the GT image for the generator is rescaled from $64 \times 64$ to $32 \times 64$ to match the HPM image dimensions.
Similar to \cite{fruhstuck2022insetgan}, we observe that GANs do not handle large areas of uniform color well and, therefore, we use the tall HPM images, where the number of the background pixels has been reduced, rather than square images with many background pixels.
We are able to do so because our dataset is filtered and it only contains the poses where the width of the person is equal to at most one half of their height. 
Finally, to help the generator understand the link between the clothing semantics and the context, we use as a joint (HPM, background) discriminator on top of the individual HPM and background discriminators which are used to increase the quality and realism of generated HPMs and backgrounds, respectively.
In summary, the generator loss for the semantic maps is defined as:
\begin{equation}
\begin{split}
    L_S^{G} = L_{adv}^{D_\mathcal{M}} + L_{adv}^{D_{B}} +  L_{adv}^{D_{MB}}  + \weight_3 L_1 + \weight_4 L_{R} + \weight_5 L_{perc} ,
\end{split}
\end{equation}
where $L_{adv}^{D_\mathcal{M}}$ and $L_{adv}^{D_{B}}$ represent the adversarial losses that come from the HPM and the blurry background image discriminators respectively; $L_{adv}^{D_{MB}}$ is the loss from an additional discriminator that evaluates the composite of the blurry background and the HPM; $L_1$ is the $l_1$ loss applied between the generated blurry background and the corresponding generated background $\Tilde{i}_c$ from the VAE; $L_{perc}$ is a standard perceptual loss commonly used in the literature for generating more detailed images~\cite{zhang2018unreasonable}; $L_{R}$ represents the path regularisation~\cite{Karras2019stylegan2} that is applied to the generator; $\weight_3$, $\weight_4$, and $\weight_5$ are the weighting factors for the specific losses.
%

%%%%%%%%%%%%%%%%%%%%%%%%%%%%%%%%%%%%%%%%%%%%%%%%%%%%%%%%%%%%

\subsection{HPM translation module}
In the remainder of the pipeline, we use diffusion models~\cite{ho2020denoising}.
In particular, we leverage the large pre-trained latent diffusion model Stable Diffusion~\cite{rombach2021highresolution} as our image prior.
Diffusion models are well-known for their capacity to generate highly detailed images.
They do this gradually, through the denoising diffusion process in $T$ timesteps.
During training, Gaussian noise $\mathcal{N}(\mathbf{0}, \mathbf{I})$ gets added to the original sample $\mathbf{x}_0$, step by step, and a UNet-like model~\cite{huang2020unet} is trained to predict the noise $\epsilon$ residuals, at timestep $t$.
For latent diffusion models (LDM), where the denoising process is performed in the latent space rather than in pixel space, $\mathbf{z}_t$ marks the corresponding latent representation of a noisy sample $\mathbf{x}_t$ at timestep $t$.
The LDM objective is then formulated as follows:
\begin{equation}
L := \expec_{\encoder(x), \epsilon \sim \mathcal{N}(0, 1),  t}\Big[ \Vert \epsilon - \model(z_{t},t) \Vert_{2}^{2}\Big] \, .
\label{eq:ldmloss}
\end{equation}
Given the large success of LDMs, new mechanisms have appeared on how to control their image generation process further.
In this paper, we use the ControlNet~\cite{zhang2023adding} as a backbone.
The main idea behind ControlNet is to create a trainable copy of a pre-trained LDM, attach it to the LDM and in turn train it without modifying the original model (zero-convolutions).
In this way, generated images retain the quality from the original LDM which remains unchanged.
The copy is additionally conditioned on images $\mathbf{c}_\text{f}$ that are encoded into its latent space by a small learnt neural network and the feature maps of the encoder of the LDM copy are added to the feature maps of the original LDM so as to transfer the original conditioning signal.
Including the text conditioning $\mathbf{c}_t$, the objective function for a conditional LDM is now equal to:
\begin{equation}\vspace{-3pt}
\begin{split} 
L^C = \mathbb{E}_{\mathbf{z}_0, \mathbf{t}, \mathbf{c}_t, \mathbf{c}_\text{f}, \epsilon \sim \mathcal{N}(0, 1) }\Big[\Vert \epsilon - \epsilon_\theta(\mathbf{z}_{t}, \mathbf{t}, \mathbf{c}_t, \mathbf{c}_\text{f})) \Vert_{2}^{2}\Big] \, .
\end{split}
\label{eq:loss}
\end{equation}
We attach a ControlNet module to the Stable Diffusion model and train it on approx. $23K$ images, where the input conditioning are RGB HPM images and the output are the corresponding images of real people in the wild.
During training, we use a generic text prompt that is dropped $50\%$ of the time.
This model is later used as a translator for our custom low-resolution RGB HPM images that are padded and then upscaled from $32 \times 64$ to $512 \times 512$ during training.
The output is upscaled from $200 \times 200$ to $512 \times 512$ during training.

\begin{figure}[t!]
\centering
    \includegraphics[width=\columnwidth]{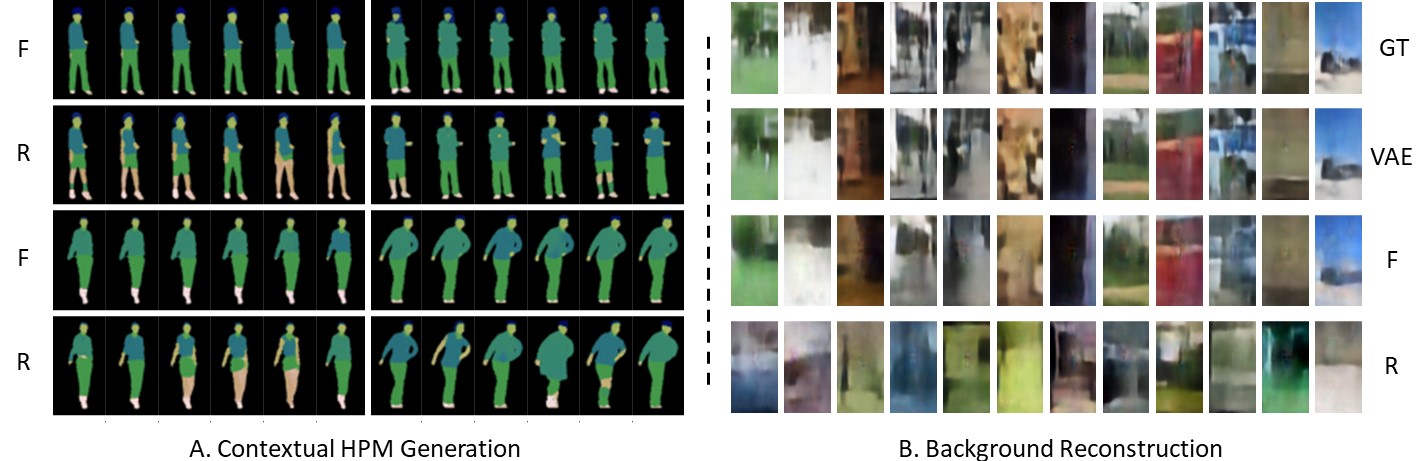}
    \caption{
    \textbf{Analysis of the StyleGAN generator:}
    \textit{A. Contextual HPM Generation:}
    Fixing the context (rows F) generates more uniform clothing
    than varying the context embeddings (rows R).
    \textbf{This indicates that there is a link between the context and the environment} that has been learned by our context-aware HPM generator. 
    \\
    \textit{B. Background Reconstruction:} 
    GT/VAE reconstructions are shown in rows 1/2; fixing the context vector in the StyleGAN generator gives matching results (row F) with GT/VAE, while varying the context embeddings leads to random predictions (row R).}
    \label{fig:hpm-background}
\end{figure}

%%%%%%%%%%%%%%%%%%%%%%%%%%%%%%%%%%%%%%%%%%%%%%%%%%%%%%%%%%%%

%
\subsection{Inpainting and super-resolution}
The human body is highly complex due to its large range of poses and also appearances, as one body may be clothed in many different types of outfits.
For this reason, there exists a large semantic gap between the body and its binary segmentation mask.
This gap makes it especially difficult to inpaint human bodies into photographs, when only the ambiguous binary mask is available and the body is to be inpainted into the specified region.
To this end, at inference time, we condition the inpainting LDM on a semantic representation of the body and clothing (HPM) via our pre-trained HPM translation module (ControlNet).
In this way, we are able to transfer the background from the original image into the resulting image using the inpainting model while predicting the HPM-compatible humans at the same time.
The impact of the HPM translator can be adjusted with a weight $\beta$ (controlNet strength) that is used for the multiplication of the ControlNet's feature maps prior to their addition with the feature maps of the inpainting model.
A low weight leads to a reduced guidance while producing higher quality imagery.
We found that setting $\beta$ to $0.5$ leads to satisfactory results.

\begin{table}[tb]
\caption{\textbf{Quantitative analysis:} C denotes whether the context vector is fixed (F) or random (R). Fixing the context vector leads to predictions that are closer to GT, especially in background reconstruction. The results are calculated over 10K contexts. The analysis confirms that our method is context-aware (PSNR, MSE, LPIPS) and generates meaningful HPMs and backgrounds (FID, KID). The performance of the VAE module used to generate the blurry contexts (rows 2 and 5) is shown in row 7.}
  \centering
    \begin{tabular}{crccccc}
    \toprule
     \textbf{C} & \textbf{Evaluation}  & PSNR $\uparrow$& MSE $\downarrow$ & LPIPS $\downarrow$ & FID $\downarrow$ & KID $\downarrow$\\
    \midrule
     & HPM  & \textbf{20.44}&\textbf{0.0102} & \textbf{0.051}  & \textbf{11.5} & \textbf{0.0098} \\
    
     F & Blurry Context & \textbf{23.33} & \textbf{0.0060} & \textbf{0.123} & \textbf{41.1} & 0.0412\\
    
     & Real Context & \textbf{21.94} & \textbf{0.0078} & \textbf{0.154} & - & - \\

    \midrule
     & HPM  & 20.37 & 0.0104 & 0.053  & 12.2 & 0.0104 \\
    
     R & Blurry Context & 11.81 & 0.0769 & 0.361 &  41.4 & \textbf{0.0389}\\
    
     & Real Context & 11.71 & 0.0781 & 0.358 & - & - \\
    
    \midrule
     F & VAE Recon.  & 27.05 & 0.0022 & 0.051 & 7.0 & 0.0029 \\
    
    \bottomrule
  \end{tabular}
\label{tab:fixed}
\end{table}

\section{Experiments}
\label{sec:experiments}

 \begin{figure}[t!]
 \centering
     \includegraphics[width=\columnwidth]{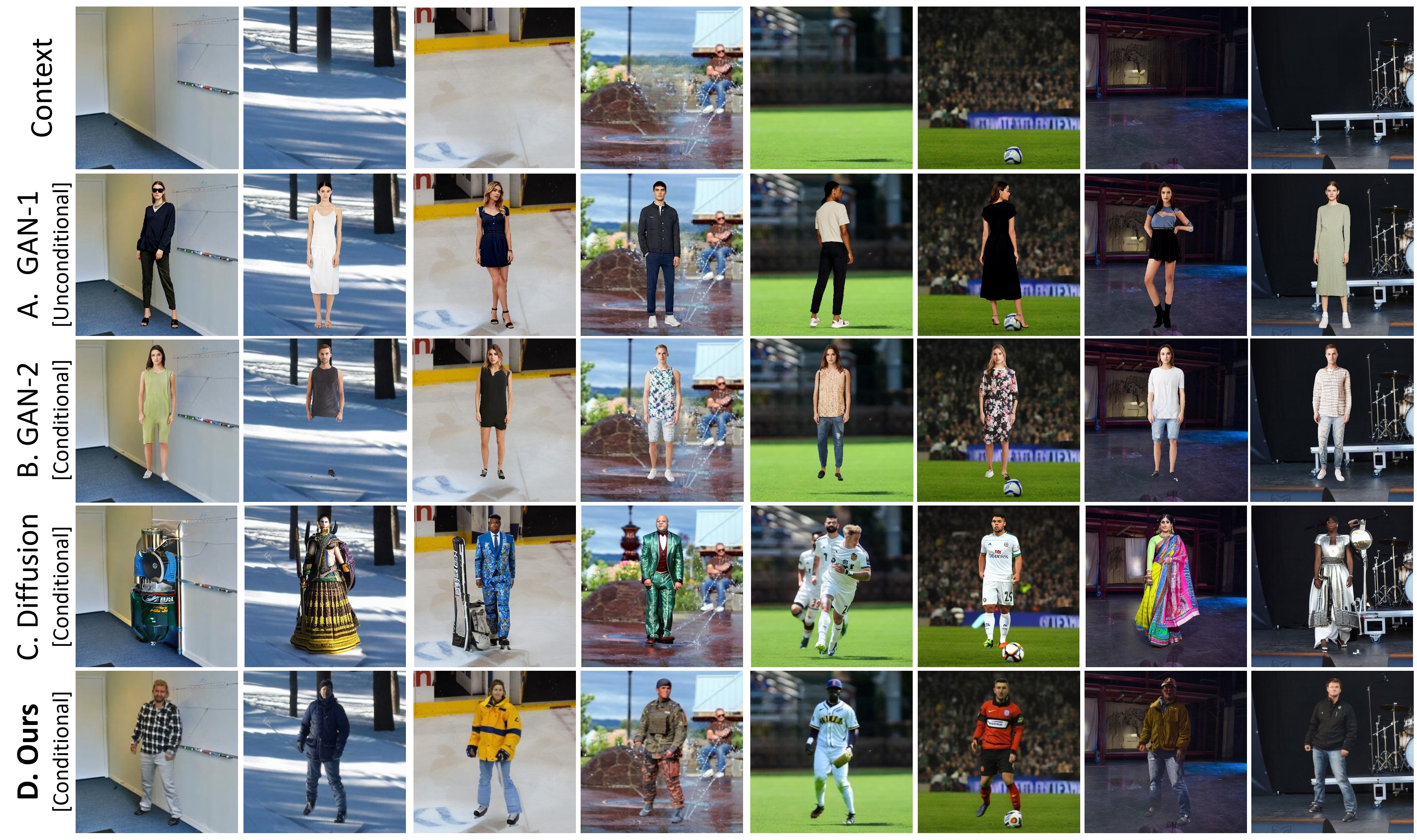}
     \caption{
     \textbf{Comparison against state of the art:}
    we compare (A) StyleGAN-Human, 
    (B) TextHuman,
    (C) ControlNet (OpenPose) \& SD-2.1-Inpainting,
    with (D) Our full method.
    (A) is unconditional, while (B) is conditioned on a dense pose. Both do not respect the context - considering clothing semantics, and lighting.
    Parts of the body may be missing (see (B), col 2).
    While (C) is conditioned on a pose, the pose is not always respected.
    It changes the original scene (e.g. by insertion of new objects; see col 3) and it does not always generate single humans (see cols 1, 5).
    In contrast, Ours (D) does not change the scene, it generates single humans that are posed correctly, and overall contextual alignment, when clothing semantics and lighting are considered, is higher than in (A), (B), and (C).
    \textit{Note:} (B), (C), and (D) have the same input pose.
    }
     \label{fig:sota}
 \end{figure}

\subsection{Data}
Our method utilizes the AHP dataset~\cite{zhou2021human} comprising context images, images with humans, machine-annotated 2D pose information, and HPMs. For training, we extract context images by using segmentation masks to remove people from the images. Skeleton images with pose data are generated using OpenPose~\cite{8765346, cao2017realtime}. During inference, only a context image and the pose information are required.

The AHP dataset is compiled from large-scale instance segmentation and detection datasets, including COCO~\cite{lin2014microsoft}, VOC~\cite{everingham2010pascal}, SBD~\cite{hariharan2011semantic}, and Objects365~\cite{shao2019objects365}. It underwent filtering and post-processing to yield 56K images with parsing maps and binary human masks, which are integral to our method. However, the raw AHP dataset's varied resolution and quality, along with inconsistent human pixel percentages, lead to sub-optimal results. Additionally, the provided parsing maps sometimes have disconnections and holes.
To mitigate these issues, we further refined the AHP images. Images were selected based on human size, with a minimum dimension of $180$ pixels in height and 90 pixels in width, and standardized to a resolution of \(200 \times 200\). We also filtered out images with mismatched binary masks and parsing maps to prevent background bleeds into human images. To address occlusions in parsing maps, we added a segmentation category for accessories and discarded images with disconnected parsing map parts.
Our final dataset consists of $\sim23$K images, each paired with its corresponding parsing map and OpenPose keypoints, ensuring higher consistency and quality for training.

%%%%%%%%%%%%%%%%%%%%%%%%%%%%%%

\begin{table}[tb]
\caption{\textbf{Truncation trick:} the table shows FID$\downarrow$ scores for varying $\psi$ across 10K randomly selected contexts. We use the optimal value of $\psi$, equal to $1.0$, in this paper.}
  \centering
    \begin{tabular}{ccccccc}
    \toprule
     \textbf{Output} & \textbf{$\psi=0.5$}  & \textbf{$\psi=0.6$} & \textbf{$\psi=0.7$} & \textbf{$\psi=0.8$} & \textbf{$\psi=0.9$} & \textbf{$\psi=1$} \\

    \midrule
     HPM & 15.73 &  14.59 & 13.53 & 12.64 & 11.96 & \textbf{11.49} \\

     Context & 112.45 &  86.33 & 67.55 & 54.77 & 46.29 & \textbf{41.08} \\
    
    \bottomrule
    \end{tabular}
\label{tab:psi}
\end{table}

%%%%%%%%%%%%%%%%%%%%%%%%%%%%%%

\begin{table}[tb]
\caption{\textbf{ControlNet strength:} the table shows FID$\downarrow$ scores for varying $\beta$ across 10K images. Our method outperforms LDM from \cite{zhang2023adding}, using the same inpainting model~\cite{rombach2021highresolution}.}
  \centering
    \begin{tabular}{cccccc}
    \toprule
     \textbf{Method}  & \textbf{$\beta=0$} & \textbf{$\beta=0.25$} & \textbf{$\beta=0.5$} & \textbf{$\beta=0.75$} & \textbf{$\beta=1$} \\
    \midrule
     Baseline & - &  - & 78.7 & 86.7 & 90.0  \\
     \textbf{Ours} & 42.3 & 22.6 & 21.0 & \textbf{19.8} & 20.2  \\
    \bottomrule
    \end{tabular}
\label{tab:beta}
\end{table}

\subsection{Comparison to state of the art}
\label{sec:sota}
\begin{figure}[t!]
\centering
    \includegraphics[width=\columnwidth]{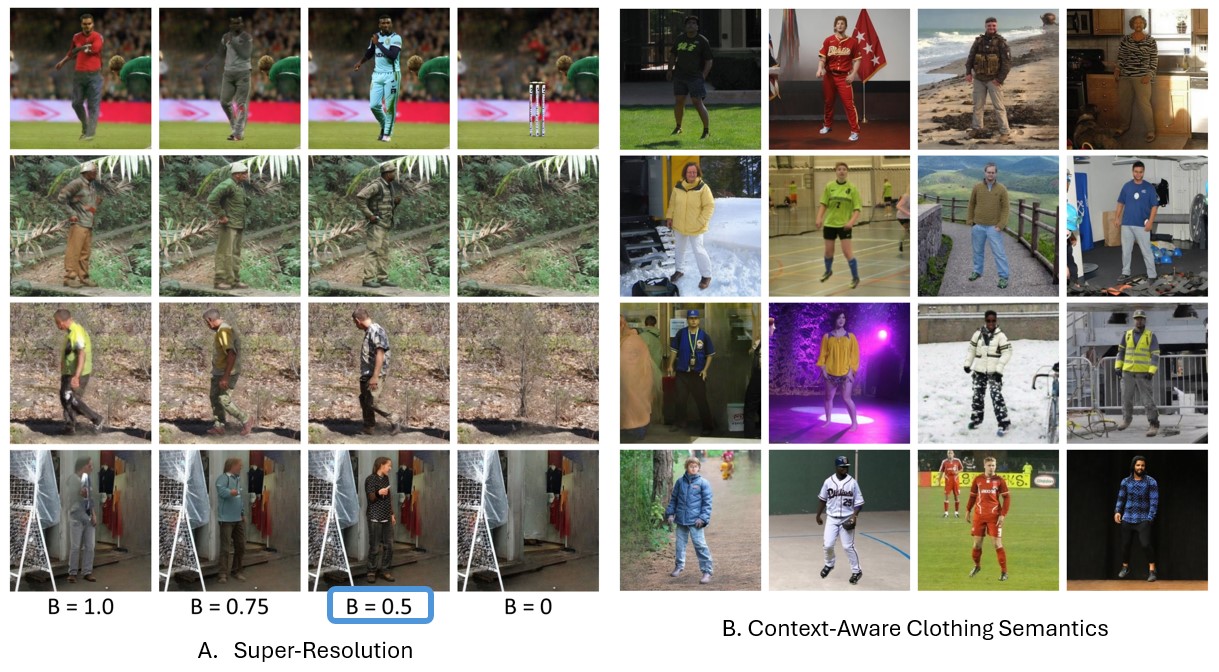}
    \caption{\textit{A. Super-Resolution:} higher image quality may be achieved by lowering the ControlNet strength via $\beta$. However, when $\beta$ is too low, humans are no longer generated ($\beta=0$ shows the vanilla SD inpainting result without additional HPM guidance).
    \textit{B. Fixed pose:} generated clothing changes depending on the context.
    }
    \label{fig:super-pose}
\end{figure}

\textbf{Qualitative/quantitative analysis:}
Figure~\ref{fig:sota} contains a comparison with state-of-the-art full-body generation methods including (1) an unconditional GAN-based method (StyleGAN-Human~\cite{fu2022stylegan}), (2) dense pose conditioned GAN-based method (TextHuman~\cite{jiang2022text2human}) as well as (3) a 2D pose conditioned diffusion-based method (ControlNet~\cite{zhang2023adding}) using the same inpainting module~\cite{rombach2021highresolution} as in Ours.
In (1), humans are generated at $1024 \times 1024$, the background is subsequently removed using rembg~\cite{rembg} and humans are then pasted on top of the context image, at random.
Next, in (2), humans are generated at $512 \times 512$ from their randomized HPM predictions, and the same procedure is followed as in (1).
In (3), rectangular masks are passed to the inpainting model as the 2D pose-conditioned ControlNet model does not provide information about the location of the body/clothing.
Humans are inpainted at $512 \times 512$.
Similarly to (3), our method inpaints humans using a diffusion model, however, we use GAN-generated HPM masks for inpainting and additional guidance.
Our method generates more natural-looking, environment-specific humans when the clothing semantics and the overall color harmonization are considered.
Table~\ref{tab:beta} shows quantitative analysis against the 2D pose-conditioned ControlNet baseline.
In Figure~\ref{fig:super-pose} (A), we gradually decrease the impact of our fine-tuned ControlNet model through $\beta$.
When $\beta=0$, our system is equivalent to the vanilla Stable Diffusion inpainting model, using a binary mask (from the predicted HPM).
It fails to inpaint humans into the photographs without additional guidance.

\begin{figure*}[t!]
\centering
    \includegraphics[width=\textwidth]{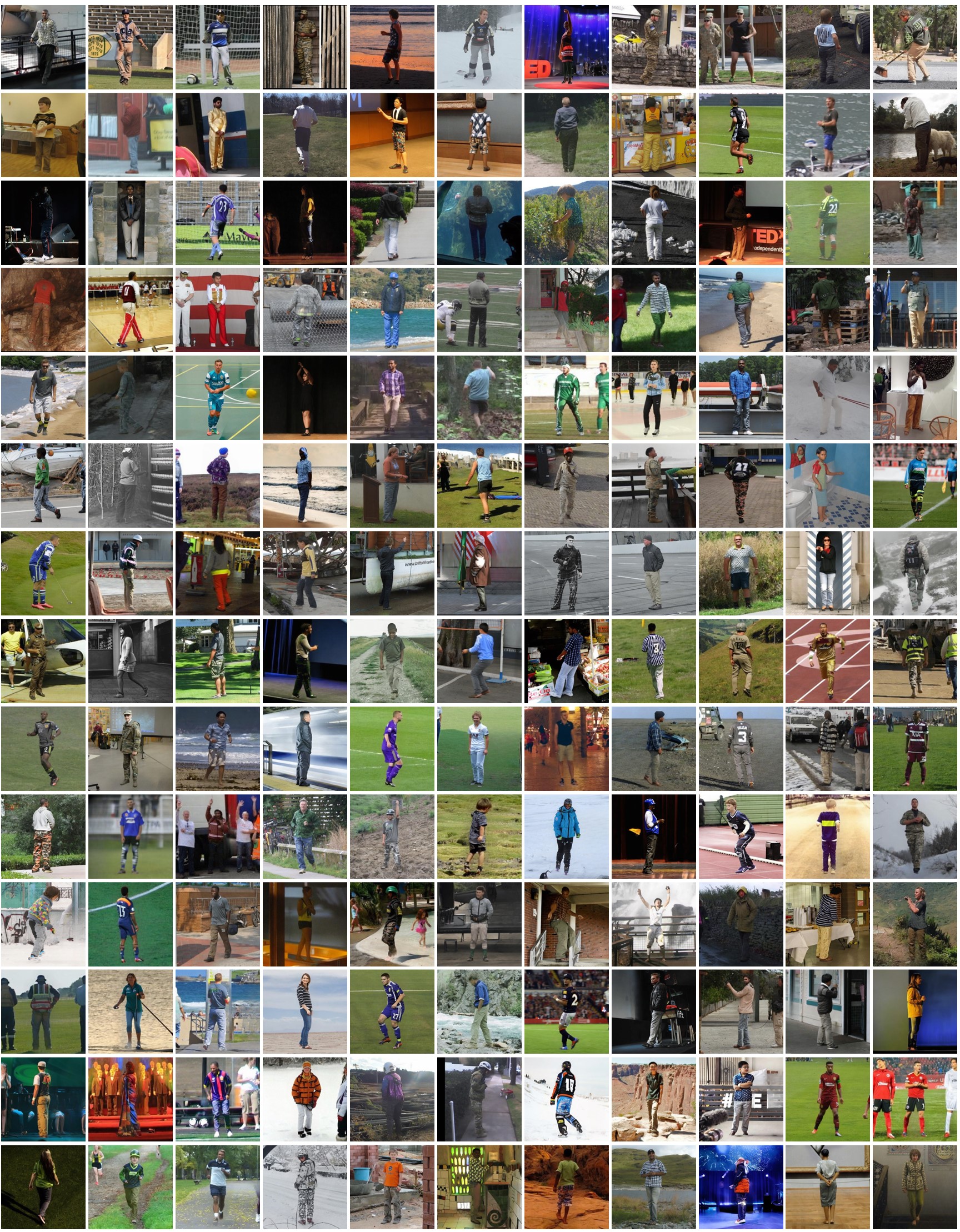}
    \caption{%
    \textit{ESP} outputs full-body humans wearing environment-inspired clothing that is semantically suitable for diverse contexts. 
    We show a wide range of different scenes including indoors and outdoors, with varying image resolutions (from $200 \times 200$ to $512 \times 512$), target pose, and lighting conditions.
    The produced results depend on the quality of the context image and on the capabilities of the generic Stable Diffusion 2.1 inpainting model that is used in the final step of our system.
    Please zoom in for details.
    }
    \label{fig:results}
\end{figure*}

\begin{figure}[t!]
    \centering
    \includegraphics[width=1\columnwidth,trim={0 0 28.1cm 0},clip]{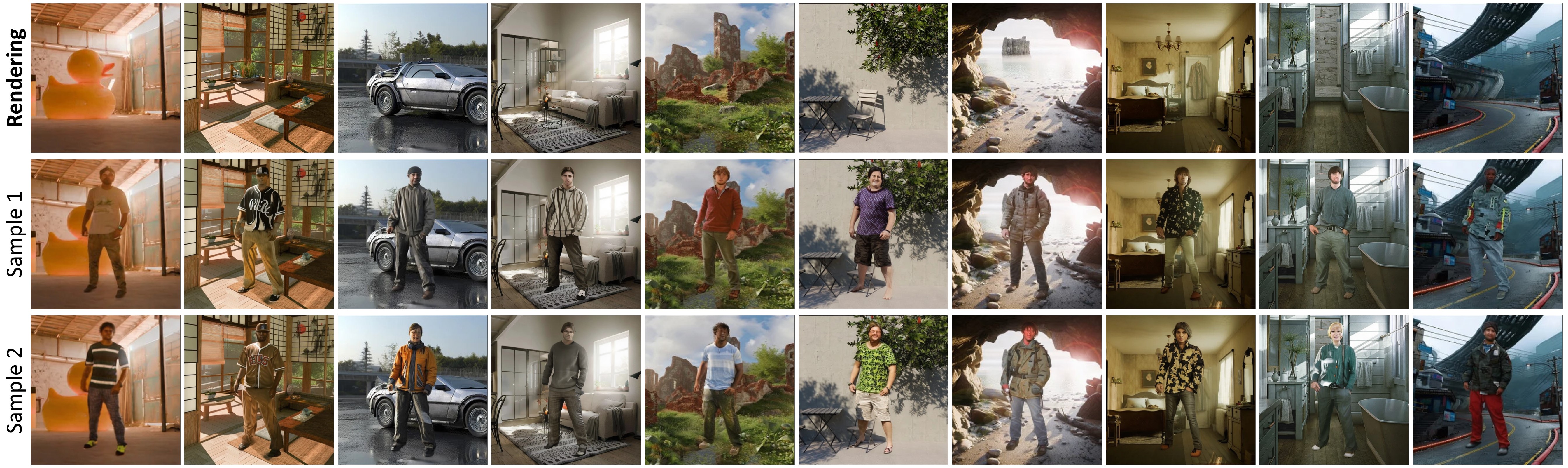}
    \caption{\textbf{Out-of-domain generalization:} our results on top of synthetic backgrounds. }
    \label{fig:synthetic}
\end{figure}

\noindent
\textbf{Perceptual evaluation:}
To evaluate the appropriateness of our generated humans versus state-of-the-art, we use BLIP-2~\cite{li2023blip} to caption a set of context images and group them by the environment type. 
We then select 20 prominent environment groups and randomly select one sample for each.
We compare against the current state-of-the-art, i.e. a pose-conditioned LDM from \cite{zhang2023adding}.
16 people attended the short study and the results show that our method was preferred 62\% of the time. % 61,875
3\% of the time, the user was undecided. %3.4375

\subsection{Evaluation and ablation studies}

\noindent
\textbf{Quantitative analysis:}
To determine the effectiveness of our HPM generator, we compare it to GT data with a fixed context C (denoted as F) and with a random context (R) in Table~\ref{tab:fixed}.
The results demonstrate that our method generates meaningful parsing masks and background images when the original data distribution is considered (FID, KID). 
It is also shown that the background reconstruction errors (PSNR, MSE, LPIPS) are considerably lower when the context vector is fixed because our method generates context-aware predictions.
The reconstructed backgrounds (row 2) are similar to the VAE predictions (row 7).
The VAE module provides embeddings that allow for the HPM model to generate backgrounds that look similar to the real ones (row 3).
Finally, Table~\ref{tab:psi} shows the ablation of the truncation trick $\psi$ parameter from ~\cite{Karras2019stylegan2}.

\noindent
\textbf{Context embeddings:}
Figure~\ref{fig:hpm-background} (B) shows the blurry background reconstructions of our (1) VAE, and (2) contextual HPM StyleGAN generator. 
StyleGAN's predictions are of lower quality than the VAE.
Nevertheless, we find that these reconstructions are meaningful as they also lead to the generation of environment-inspired clothing semantics.
In Figure~\ref{fig:hpm-background} (A), the difference between the fixed and random context embeddings is shown in the generator. 
Fixing the context (rows F) generates more uniform clothing.
Here, each sample is generated with a different random part of the contextual style vector, while the context embeddings are fixed.
On the contrary, when the context embeddings are not fixed, our model generates more varied clothing styles (rows R). 
This indicates there is a link between the context and the environment.

\noindent
\textbf{Inpainting:}
Changing the impact of our ControlNet model $\beta$ may increase image quality (Figure~\ref{fig:super-pose} (A)).
This is due to the low resolution of images that we use to train the model.
When $\beta$ is lowered, the original Stable Diffusion model that was trained on high-quality images has more impact on the final result, and therefore images of higher quality are generated.
However, if $\beta$ is too low, generated humans might no longer respect the input pose or they might not be generated, as discussed above in Sec.~\ref{sec:sota}.
For numerical analysis, see Table~\ref{tab:beta}.

\noindent
\textbf{Pose conditioning:}
Previously, it had been shown that pose plays an important role in the synthesis of scene-compatible humans~\cite{brooks2022hallucinating}.
Here, we argue that the clothing semantics may also be inferred from the scene context alone.
We fix the pose in Fig.~\ref{fig:super-pose}(B) and Fig.~\ref{fig:synthetic} to show how our method generates plausible clothing.
More examples with the original pose are shown in Fig.~\ref{fig:teaser} and Fig.~\ref{fig:results}.

\begin{figure*}[t!]
\centering
    \includegraphics[width=0.9\textwidth]{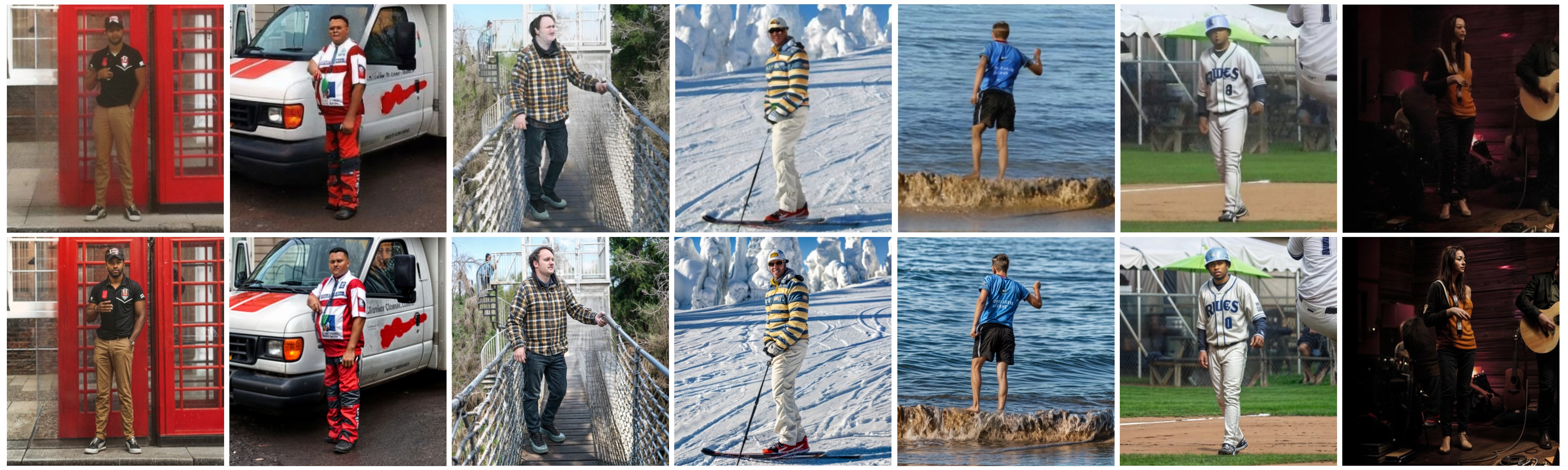}
    \caption{
    \textbf{Upscaling:}
    the results of our method (top row) may be further improved using an upscaler (bottom row)~\cite{upscale}. 
    }
    \label{fig:image-quality}
\end{figure*}

\noindent
\textbf{Limitations:}
Our method generates people in simple poses only (see Sec.~\ref{sec:experiments}), and, if the conditioning pose is not plausible, the method might fail to generate a valid HPM.
Also, the quality of our results depends on the capabilities of the generic inpainting model and may be further enhanced using an upscaler (Fig.~\ref{fig:image-quality}).

\noindent
\textbf{Potential negative impact:}
Our system is biased as there is a lack of racial diversity in specific contexts, and the gender imbalances skewed toward male representations are a reflection of the training data's limitations.
Therefore, it is imperative to take conscious steps to mitigate these biases.
This can include diversifying the training data which is out of the scope of this paper.

\section{Conclusion}
\label{sec:conclusion}

We have demonstrated how full-body human generation and subsequent inpainting can be made context-aware, i.e., our generated humans are in sync with the environment when their clothing semantics and overall color harmonization are considered.
Our main contribution is conditioning the image generation process on scene context to generate environment-specific people in clothing that is suitable for inpainting into a given environment.
Our results demonstrate that our method produces plausible contextualized people in scenes, outperforming state-of-the-art methods on the task of contextual full-body generation.
\noindent \textbf{Acknowledgements:}
We thank Soubhik Sanyal for his assistance in writing, and Benjamin Pellkofer for IT support.
This work has received funding from the European Union’s Horizon 2020 research and innovation programme under the Marie Skłodowska Curie grant agreement No 860768 (CLIPE project).

\noindent \textbf{Disclosure:}
\href{https://files.is.tue.mpg.de/black/CoI_ECCV_2024.txt}{https://files.is.tue.mpg.de/black/CoI\_ECCV\_2024.txt}
\clearpage

% ---- Bibliography ----
\bibliographystyle{splncs04}
\bibliography{main}
\end{document}